\documentclass[runningheads]{llncs}
\usepackage[T1]{fontenc}
\usepackage{graphicx}
\usepackage{times}
\usepackage{helvet}
\usepackage{courier}
\usepackage{amsmath}
\usepackage{algorithm}
\usepackage{algorithmic}
\usepackage{csquotes} 
\usepackage{color}
\usepackage{paralist}
\usepackage{amssymb}
\usepackage{indentfirst}
\usepackage{subfigure}
\usepackage{float}
\usepackage{multirow}
\usepackage{cite}
\usepackage{mathrsfs}

\usepackage{mathrsfs} 
\usepackage{amsfonts}
\usepackage{todonotes}
\usepackage{pgfplots} 
\usepackage{epstopdf}
\usepackage{epsfig}
\pgfplotsset{compat=newest}
\usepackage{color}
\definecolor{forestgreen}{RGB}{0,139,69}
\usepackage{caption}

\usepackage[switch]{lineno}

\usepackage[colorlinks, linkcolor=red,  anchorcolor=blue, citecolor=blue]{hyperref}
\usepackage{textcomp,booktabs}
\usepackage{amssymb}
\usepackage{pifont}
\newcommand{\cmark}{\ding{51}}%

\usepackage{makecell}

\begin{document}
\title{ Multi-modal Fusion based Q-distribution Prediction for Controlled Nuclear Fusion 
\thanks{Corresponding author: Yifeng Wang (\email{yfwang@ipp.ac.cn})}   
}
%


\author{
Shiao Wang \inst{1} \and 
Yifeng Wang \inst{2} \and 
Qingchuan Ma \inst{1} \and 
Xiao Wang \inst{1} \and 
Ning Yan \inst{2} \and \\ 
Qingquan Yang \inst{2} \and 
Guosheng Xu \inst{2} \and 
Jin Tang \inst{1}
}

\authorrunning{Shiao Wang et al.}
%
\institute{
School of Computer Science and Technology, Anhui University, Hefei, China \and 
Institute of Plasma Physics, Chinese Academy of Sciences, Hefei, China  
}


\maketitle              
\begin{abstract}
Q-distribution prediction is a crucial research direction in controlled nuclear fusion, with deep learning emerging as a key approach to solving prediction challenges. In this paper, we leverage deep learning techniques to tackle the complexities of Q-distribution prediction. Specifically, we explore multimodal fusion methods in computer vision, integrating 2D line image data with the original 1D data to form a bimodal input. Additionally, we employ the Transformer's attention mechanism for feature extraction and the interactive fusion of bimodal information. Extensive experiments validate the effectiveness of our approach, significantly reducing prediction errors in Q-distribution.
\end{abstract}

\section{Introduction} \label{sec:introduction} 
The problem of nuclear fusion has always been a major challenge that humanity needs to solve. In recent years, many researchers have made outstanding contributions to the development of controllable nuclear fusion~\cite{lennholm2024plasma, seo2024avoiding, lennholm2024controlling}. Research on international magnetic confinement controlled nuclear fusion began in the 1950s. Over the years, various approaches have been explored, including magnetic confinement~\cite{dal2024measurement, wang2023current}, magnetic mirrors~\cite{van2024modulated, velasco2023robust}, stellarators~\cite{thienpondt2023prevention, goodman2024quasi, rodriguez2023trapped}, and Tokamaks~\cite{zheng2023disruption, schwander2023global, berkery2024nstx}, all aimed at improving key plasma parameters to achieve the conditions necessary for controlled nuclear fusion reactions eventually. Since the 1970s, the Tokamak method has gradually demonstrated unique advantages, becoming the mainstream approach in magnetic confinement fusion research. With the global advancement of Tokamak experiments, the comprehensive plasma parameters have been continuously improved, and significant progress has been made in fusion engineering technologies. However, many key technologies still face considerable challenges before practical applications can be realized.

Among the many sub-tasks in controlled nuclear fusion research, Q-distribution prediction is both a critical and highly challenging problem. In this paper, we leverage artificial intelligence techniques to address the Q-distribution prediction challenge within the context of controlled nuclear fusion. To begin, we create a dataset that contains 5,753 data samples using a simulation toolkit. For efficient model training and evaluation, these data are divided into 5,166 samples for training and 587 for testing. Based on this dataset, we developed a deep learning model specifically designed for Q-distribution prediction. Our approach focuses on exploring multimodal fusion techniques within the field of computer vision. To capture temporal trends in the data, we transformed the original 1D data into 2D line charts, visually representing how key indicators change over time. We then employed the Vision Transformer (ViT)~\cite{dosovitskiy2020vit}, a state-of-the-art model for image analysis, as the core of our multimodal fusion network. This network efficiently extracts and fuses features from the multimodal inputs. Finally, the model performs regression to predict the Q-distribution and assess its degree of alignment with the ground truth. The results of our approach demonstrate the efficacy of using deep learning for this task, offering a significant improvement in prediction accuracy over previous methods.


Overall, we have made the following contributions to this work:

1). We propose a multimodal fusion framework for Q-distribution prediction using convolutional neural networks, multi-layer perceptron, and Transformer networks. It is the first multimodal framework for the Q-distribution prediction.  

2). We simulate a new dataset that contains 5753 pieces of data for the Q-distribution prediction and divide them into training and testing sets to train our network. 

3). We conduct extensive experiments on this dataset, which fully demonstrate the effectiveness of our proposed multimodal fusion method for Q-distribution prediction.

\section{Method} \label{sec:method}

\subsection{Problem Formulation} \label{sec:probFormulation}
\begin{figure*}
    \centering
    \includegraphics[width=4.8in]{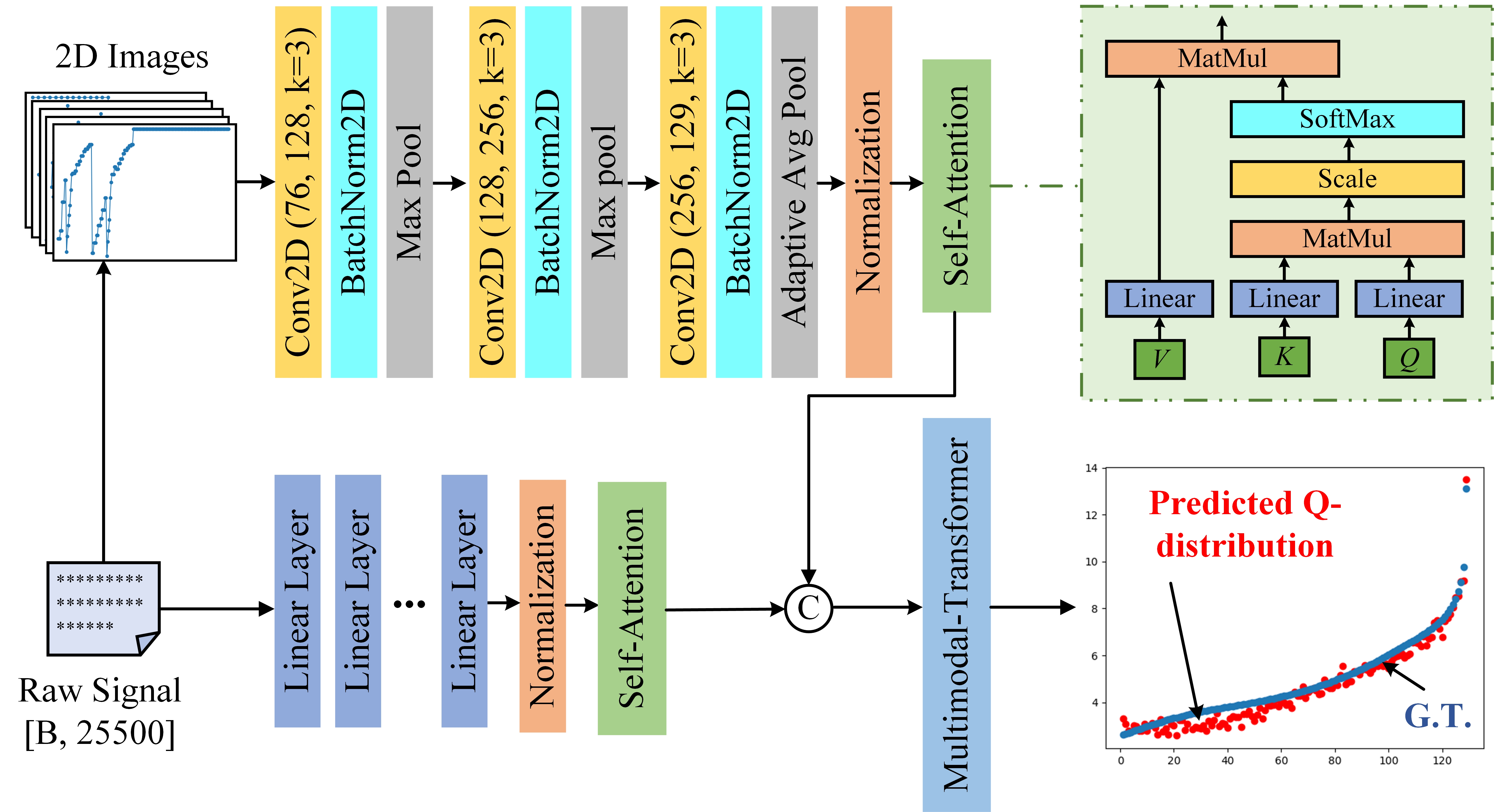}
    \caption{An overview of our proposed model for the Q-distribution.} 
    \label{fig:framework}
\end{figure*}

In the task of predicting plasma ruptures, accurately forecasting the Q-distribution is of critical importance. This paper focuses on applying deep learning methods to minimize errors in Q-distribution prediction. Selecting an appropriate deep learning approach for modeling this unique type of nuclear fusion data is crucial. Moreover, from a multimodal artificial intelligence perspective, it is also essential to explore how to integrate information from multiple viewpoints to enhance model accuracy. Consequently, we conducted comprehensive experiments with various mainstream deep learning models and identified the optimal data fusion method to achieve effective Q-distribution prediction. We design a simple and effective multimodal fusion framework for Q-distribution prediction, as shown in Fig.~\ref{fig:framework}.

\subsection{Overview} 
Deep learning has garnered significant attention in recent years due to its remarkable capabilities in various domains. In this paper, we delve into the practical implications of utilizing deep learning methodologies for nuclear fusion prediction in physical applications. To maximize the accuracy of prediction performance, our emphasis lies in leveraging multimodal deep learning network frameworks such as Transformer ~\cite{vaswani2017attention} and convolutional neural networks(CNNs)~\cite{LeCun1998GradientbasedLA}. Specifically, we begin by generating line charts for each nuclear fusion indicator, illustrating their respective changes over time as visual information. This bimodal data approach proves effective in capturing nuanced patterns. Subsequently, we employ linear network layers and convolutional neural networks to extract feature information of the raw numerical data and visual representations, respectively. Finally, we fuse the extracted features from both modalities, enhancing the overall predictive capability. Our network framework diagram is shown in Fig.~\ref{fig:framework}. To be specific, we leverage multiple linear network layers and Attention mechanisms~\cite{vaswani2017attention} to extract original features from the raw fusion index data. Concurrently, for the newly generated image data, Convolutional Neural Networks (CNNs) are employed alongside Attention mechanisms to extract visual features. The utilization of Attention mechanisms significantly enriches the expression of features, owing to its outstanding performance. Subsequently, we harness the formidable capabilities of multimodal Transformer blocks to effectively fuse the features extracted from both modalities, culminating in the final prediction. Further elaboration on the network architecture and the application of deep learning-based methods in nuclear fusion prediction will be provided in the subsequent subsections.

\subsection{Input Processing}  
In this study, our nuclear fusion raw data comprises 141 physical indicators, denoted as $\mathcal{R}$, which encompass variables such as $npsip-q$ and $q$ distribution. Notably, with the exception of $npsip-q$ and $q$, all other indicators signify the plasma rupture status during the nuclear fusion process (The $npsip-q$ and $q$ distribution represent the final predicted ground truth). To facilitate the network's interpretation and processing, we transform the content of each indicator in the raw data into tensor representations. Subsequently, we concatenate these tensors along the first dimension and feed them into the network.

In addition to the raw data for nuclear fusion prediction, we have also generated 2D image data illustrating changes in translation data indicators. Specifically, to address the varying values and scales of the raw data, we selected 76 representative predictive indicators and transformed them into 2D visual line charts.  For the convenience of drawing and calculation, we established a standard sampling approach: for each indicator of the raw data, if the number of values contained in the indicator $N$ is greater than $M$, we sample $M$ values at intervals; otherwise, we retain the original $N$ values. Based on this standard, we have drawn line charts for each representative indicator. Subsequently, we input both the raw data and the generated 2D visual data into multiple linear network layers and convolutional neural networks (CNNs) respectively for feature extraction. The detailed network structures will be introduced in the next section.


\subsection{Network Architecture} 
We have devised a network framework for nuclear fusion prediction tasks based on the multimodal fusion concept of deep learning, as illustrated in Fig.~\ref{fig:framework}. The framework primarily comprises two branches, each dedicated to processing raw data and generated image data, respectively. The detailed process is outlined below.

\noindent \textbf{Raw data processing.~} 
In order to achieve efficient expression of data, the input size of our original data is $\mathbb{R}^{B \times n}$, where $B$ represents the batch size, which denotes the amount of data loaded for processing in a single iteration, and $n$ represents the size of each individual data point, which is usually 25500. We input a batch of data into the neural network, as shown in the lower part of Fig.~\ref{fig:framework}. We begin by feeding the raw signal data into a backbone network comprising several consecutive linear network layers interspersed with sigmoid activation functions, forming what is commonly known as a Multi-Layer Perceptron (MLP), to extract the raw data features. Subsequently, we employ an Attention network to establish global dependencies among the data points, the network structure diagram of the Attention network is shown in the top right corner of Fig.~\ref{fig:framework}. We transform the raw data into Query (Q), Key (K), and Value (V) vectors, which are then fed into the linear layer within the Attention mechanism for weighting. Following this, we perform matrix multiplication between the Q and K matrices, which are subsequently normalized using a scaling factor. Finally, applying the softmax activation function to these normalized values, we obtain attention weights, which are then used to dot multiply with the Value vectors (V), yielding the final output denoted as $output_1$. The processing formula for the raw data can be summarized as follows,
\begin{equation}
    \label{Attention} 
output1 = Attention(Norm((MLP(R_i))*N))
\end{equation}

\begin{equation}
    \label{rawdata_processing} 
Attention(Q, K, V) = softmax(\frac{QK^T}{\sqrt{d_k}})V
\end{equation}
where $R_i$ represents the input of the raw data, $MLP$ represents the multi-layer perception, Norm represents the linear normalization layer, $N$ represents The number of $MLP$ layers, $Q$, $K$, and $V$ are query, key, and value matrices separately, $d_k$ in the $Attention$ formula represents the scaling factor, and $softmax$ represents the normalization operation.

\noindent \textbf{The generated 2D images processing.~}
Multi-modal fusion represents a pivotal research avenue in deep learning. This paper proposes a fusion prediction approach leveraging multimodal fusion technology. To integrate multimodal data input, we generated 2D images derived from the raw data to encompass visual information. After obtaining the 2D images that reflect the changes in indicators, we input the corresponding images into the upper part shown in Fig.~\ref{fig:framework}. Firstly, we resize the input 2D images to a predetermined fixed resolution size, which is $224 * 224$ used in this paper. Following the resizing step, we overlay a batch of data images along the first dimension to construct a $4-dimensional$ visual image dataset, which is $\mathbb{R}^{B \times K \times W \times H}$. Here $B$ refers to batch size, $K$ refers to the number of images generated by a single data point, $H$ and $W$ refer to the height and width of the images. We proceed by feeding the preprocessed visual data into the visual backbone network.  Our visual backbone network primarily comprises convolutional neural networks (CNNs).  More precisely, we employ 2D CNNs integrated with BatchNorm layers, followed by a max pooling layer aimed at gradually reducing the resolution size of the images.  It's noteworthy that the pooling layer utilized in the final stage is the global pooling layer, which considers the global information of the images. Following the extraction of visual features, we further utilize the Attention network to encapsulate the global information of visual images, thereby enhancing the feature representation capabilities. Ultimately, the output obtained from the Attention mechanism serves as the final output of the visual branch, which we call $output_2$. The processing formula for vision data is as follows,
\begin{equation}
    \label{visiondata_processing} 
output2 = Attention(AvgPool((MaxPool(BN(Conv2D(V_i))))*N))
\end{equation}
where $V_i$ is the generated 2D images from raw data, $N$ is the number of convolutional layers, and $BN$ denotes the BatchNorm layer.

\noindent \textbf{The multimodal fusion network.~}
To leverage multimodal fusion methods for enhancing prediction accuracy and reducing prediction errors, we combine the features extracted from the raw data with those obtained from the generated 2D images. Drawing from the remarkable success of the Transformer architecture, this paper employs a multimodal Transformer approach to accomplish effective multimodal fusion. Specifically, based on the $output_1$ and $output_2$ we obtained above, we first concatenate the two outputs, and then input it to the multimodal Transformer for feature interaction and fusion. Firstly, we feed the fused features into the multi-head self-attention network to facilitate feature interaction and enhancement. Subsequently, the output from this attention mechanism is passed through a linear network layer and a feedforward network. Notably, at each step, the output is augmented by adding it to the input, thereby maximizing the preservation of the original data features. After traversing through N layers of Transformer blocks, we obtain the final fused interaction output. The formula is as follows,
 
\begin{equation}
    \label{fusion} 
output = Transformer([output1,output2])
\end{equation}
where [·] represents the concatenation operation, and $Transformer$ is the multimodal Transformer. These outputs are then utilized to calculate the loss, which is computed based on the Q-distribution of the raw groundtruth. The detailed calculation process of the loss is delineated in the subsequent section.

\subsection{Loss Function}  
Through multimodal Transformers, we obtained the final output of the network. To reflect the error between our predicted results and the groundtruth, we use the $MSE$(mean squared error) loss function to measure it. The specific loss function can be represented as,
\begin{equation}
    \label{MSEloss} 
MSE = \frac{1}{n} \sum_{i=1}^{n} (GT_i - Q_i)^2
\end{equation}
where $GT_i$ represents the groundtruth of the i-th sample, and $Q_i$ represents the distribution predicted result of the i-th sample.

\section{Experiments} \label{sec:experiments}

\subsection{Dataset and Evaluation Metric} 

The data for nuclear fusion is split into training/testing subsets which contain 5177 and 575 raw data, respectively. The MSE (Mean Squared Error) metric is adopted to measure the distance between our predicted Q-distribution and the groundtruth.

\subsection{Implementation Details} 
The training of our network is conducted end-to-end. We train our network with multimodal inputs for 130 epochs. The learning rate is 0.001, and the batch size is 16. 
We opt for Stochastic Gradient Descent (SGD)~\cite{Bottou2012StochasticGD} as the optimizer.
Our code is implemented using Python based on PyTorch~\cite{paszke2019pytorch} framework and the experiments are conducted on a server with CPU Intel(R) Xeon(R) Gold 5318Y CPU @2.10GHz and GPU RTX3090 with 24 GB memory. 

\subsection{Comparison with Other Models}  
In order to effectively extract visual features from the generated 2D images, we compared various mainstream visual backbones to identify the optimal method. Table~\ref{comparison} summarizes our findings. We evaluated several architectures including Residual Networks (ResNet18, ResNet34, ResNet50)~\cite{He2015DeepRL}, Multi-Layer Perception networks (MLP), Vision Transformer networks (ViT-B, ViT-L, ViT-H)~\cite{dosovitskiy2020vit}, VGGNet~\cite{Simonyan2014VeryDC}(VGG16, VGG19), and Ours Convolutional Neural Networks (CNNs). Through comprehensive experimental comparisons, we observed that utilizing convolutional neural networks as our visual backbone network yielded the best performance. This can be attributed to the local perception ability and strong generalization capability inherent in convolutional neural networks.
 
\begin{table}
\center
\small     
\caption{Comparison with Other Models.} 
\label{comparison}
\setlength{\tabcolsep}{10pt} 
\begin{tabular}{c|ccccc}
\hline 
\textbf{Algorithm}  & \textbf{ResNet18}    &\textbf{ResNet34}  &\textbf{ResNet50}      &\textbf{MLP} & \textbf{ViT-B}   \\
\hline 
\textbf{MSE}        &0.1037          &0.1239         &0.0768         & 0.0748         &0.0759    \\
\hline 
\textbf{Algorithm}      &\textbf{ViT-L} &\textbf{ViT-H} &\textbf{VGG16}   &\textbf{VGG19}   &\textbf{Ours}   \\
\hline 
\textbf{MSE}        &0.0743         &0.0697         &0.0746         &0.0759         &0.0696            \\
\hline 
\end{tabular}
\end{table}	
\noindent

\subsection{Ablation Study}  

\noindent \textbf{Component Analysis.~} 
This paper is about the application of deep learning in the field of nuclear fusion prediction. Our experiments have shown that incorporating deep learning into network models can indeed reduce the prediction error of Q-distribution. As shown in Table~\ref{CAResults}, what we can see is that using only the raw data through some linear network layers yields an unsatisfactory result, with an $MSE$ loss value of 0.1125. However, when we leverage the powerful modeling ability of the Attention network, the prediction loss is significantly reduced to 0.0884. Even more surprisingly, inspired by the 2D image data from computer vision, when we generated 2D image data from the raw data and added it for appropriate fusion with the raw data, the result further improved to 0.0696. These results fully demonstrate that the prediction results of nuclear fusion have been improved after the addition of the multimodal fusion network for deep learning.

\begin{table}
\center
\small     
\caption{Component Analysis results.} 
\label{CAResults} 
\setlength{\tabcolsep}{12pt} 
\begin{tabular}{c|ccc|cc} 		
\hline 
\textbf{No.}  & \textbf{MLP}  &\textbf{Attention}  &\textbf{Multi-modal fusion}  &\textbf{MSE}   \\
\hline 
1 &\cmark   &          &         &0.1125      \\
2 &\cmark   &\cmark    &         &0.0884      \\
3 &\cmark   &\cmark    &\cmark   &0.0696      \\
\hline
\end{tabular} 
\end{table}

\noindent \textbf{Analysis of the Maximum Number of Spaced Sampling Points of Images.~} 
According to the above, when we added visual information, the prediction results were further improved. However, due to the fact that some indicators in the raw data are composed of thousands of values, it is difficult to generate 2D images representing the numerical changes of the indicators. Based on past experience, we conducted interval sampling on indicators with excessive values and specified the collection of only $M$ points when the numerical value $N$ of the indicator is much greater than $M$. Thus, what is the most suitable value for $M$? We conducted experiments on the maximum interval point sampling data volume as shown in Table~\ref{points}. It is obvious that when we set the maximum number of points for interval sampling to 100, the $MSE$ error reaches its minimum. When the number of sampling points is too small, the loss of unsampled data can significantly increase the prediction error. On the other hand, when there are too many sampling points, the dense clustering can cause indistinguishable points in the generated image, also reducing prediction accuracy. Therefore, selecting an appropriate number of samples is critically important. 

\begin{table}
\center
\small     
\caption{Ablation studies on the maximum number of spaced sampling points of images.} 
\label{points}
\setlength{\tabcolsep}{12pt} 
\begin{tabular}{c|cccc}
\hline 
\textbf{Sample Points}  &\textbf{50}  &\textbf{100}  &\textbf{200}      &\textbf{500}   \\
\hline 
\textbf{MSE}             & 0.0812       & 0.0696            & 0.0785            & 0.0772         \\
\hline 
\end{tabular}
\end{table}


\begin{figure*}
    \centering
    \includegraphics[width=4in]{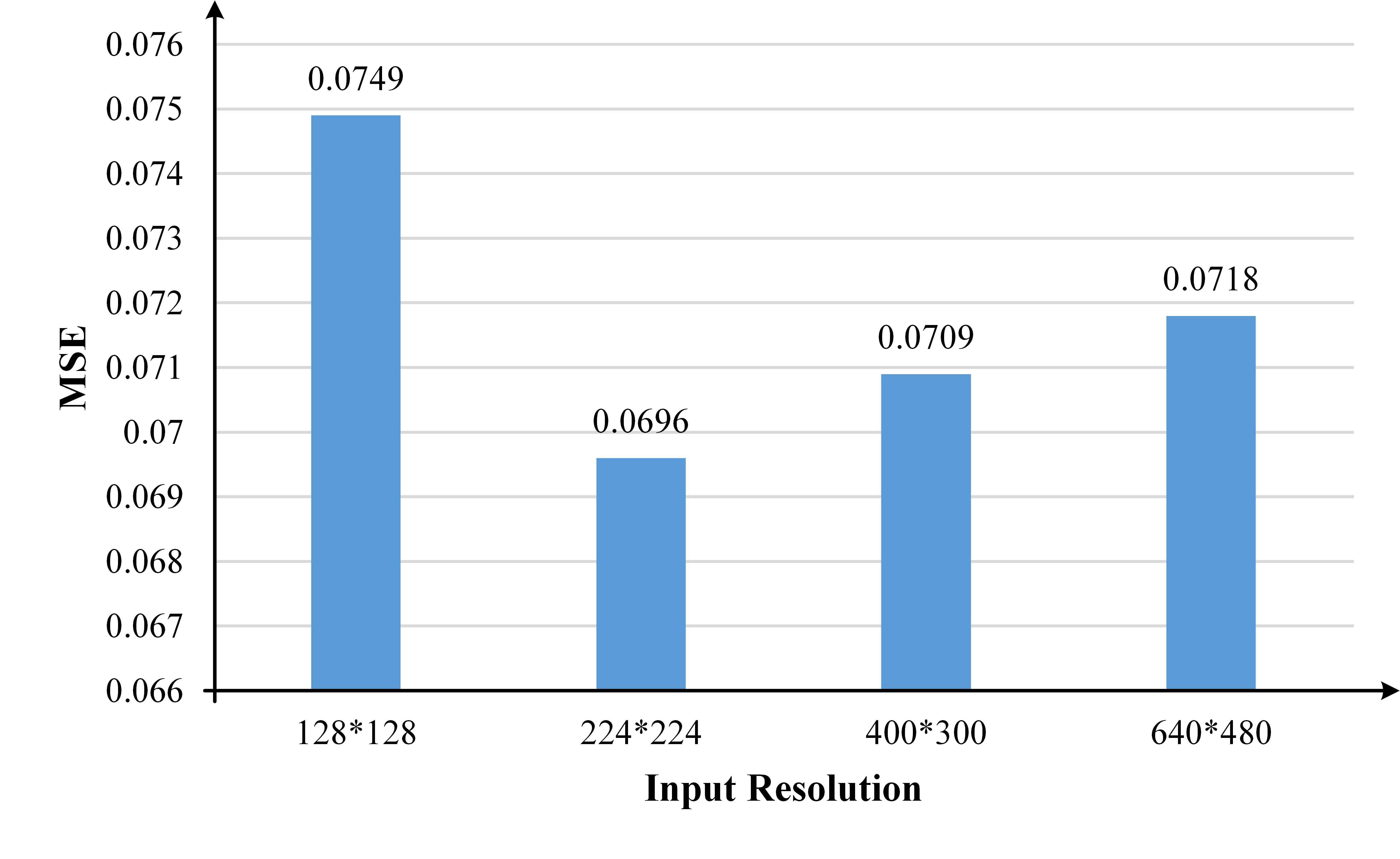}
    \caption{Visual comparison of different resolutions of input 2D images.} 
    \label{fig:resolution}
\end{figure*}

\noindent \textbf{Analysis of the Scale of 2D images.~} 
We have analyzed the maximum number of spaced sampling points of the images above. In addition, the resolution of the input images is also an important factor affecting the results. So, we set the input of images with different resolutions to find the most suitable resolution input, and the experimental results are shown in Fig.~\ref{fig:resolution}. We can observe that the best result is achieved when we choose an image resolution of 224 * 224 for input, with an error reduction of 0.0696.

\begin{figure*}
    \centering
    \includegraphics[width=4in]{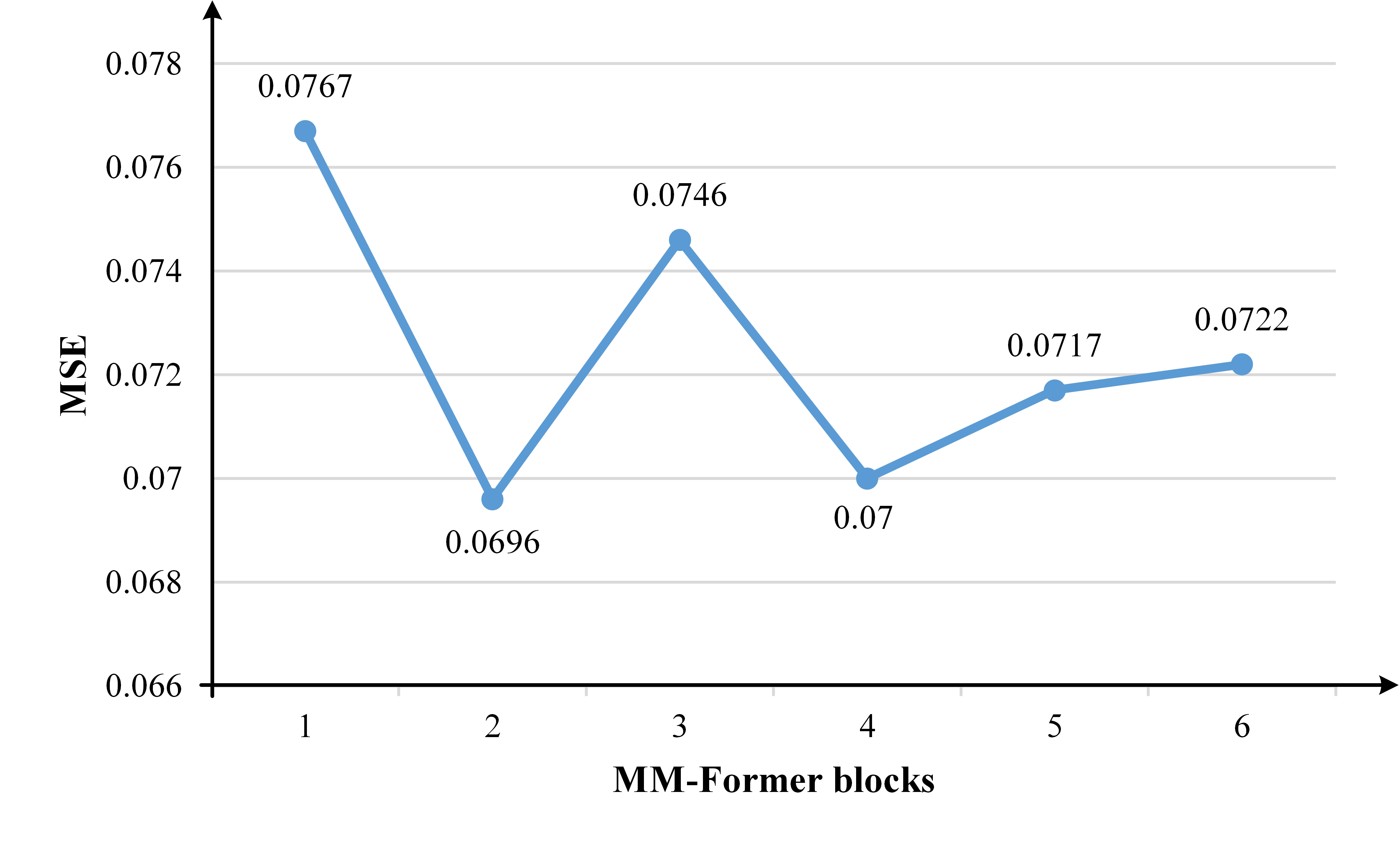}
    \caption{Visual comparison of the number of different multimodal Transformer blocks.} 
    \label{fig:MM-Formers}
\end{figure*}

\noindent \textbf{Analysis of the Number of Multi-modal Transformer Blocks.~}
Inspired by the idea of multimodal fusion in the field of computer vision, we also use multimodal Transformers to achieve multimodal fusion between raw data and generated image data. The network structure of the Transformer mainly consists of multiple Transformer blocks, each of which contains modules such as multi-head attention. In order to investigate the impact of the number of Transformer blocks on the final result, we also conduct research on the number of multi-modal transformer blocks, The experimental results are shown in Fig.~\ref{fig:MM-Formers}. Through this set of experiments, We can clearly see that the number of blocks does not directly correlate with the quality of the effect. Using only two blocks yields the optimal effect.

\section{Conclusion} \label{sec:conclusion}
In this paper, we primarily investigate how to use deep learning models to improve the predictive ability of the Q-distribution task in nuclear fusion from the perspective of multimodal fusion. Our initial attempts using simple neural networks (MLPs) to process raw data yielded suboptimal results. Consequently, we transformed the raw data into 2D visual images and leveraged Convolutional Neural Networks (CNNs) along with Attention mechanisms to effectively model the 2D image information. We then employed a multimodal Transformer network to seamlessly integrate the 1D raw information with the 2D image data, significantly reducing prediction errors in Q-distribution. Extensive analytical experiments have corroborated the efficacy of our proposed multimodal fusion framework, demonstrating its capability to effectively control Q-distribution errors in nuclear fusion. Looking ahead, we aim to further explore the potential of deep learning models in addressing various challenges in nuclear fusion tasks, including Q-distribution prediction, thereby contributing to sustainable energy development.

\small{ 
\bibliographystyle{IEEEtran}
\bibliography{reference}
}

%
%
%
%




\end{document}